\documentclass[10pt,twocolumn,letterpaper]{article}
\usepackage{arydshln}
\usepackage{cvpr}
\usepackage{times}
\usepackage{epsfig}
\usepackage{graphicx}
\usepackage{amsmath}
\usepackage{amssymb}
\usepackage{bbm}
\usepackage{float}
\usepackage{wrapfig}
\usepackage{capt-of}
\usepackage{etoolbox}
\usepackage{booktabs}
\usepackage{multicol}
\usepackage{pifont}

\usepackage[pagebackref=true,breaklinks=true,letterpaper=true,colorlinks,bookmarks=false]{hyperref}

\def\ours{\textsc{ancor}}
\def\oursspace{\textsc{ancor} }
\def\oursfull{Angular Normalized COntrastive Regularization }

\def\ourstask{C2FS}
\def\ourstaskspace{C2FS }
\def\ourstaskfull{Coarse-to-Fine Few-Shot }

\newcommand{\cmark}{\ding{51}}%
\newcommand{\xmark}{\ding{55}}%

\cvprfinalcopy % *** Uncomment this line for the final submission

\def\cvprPaperID{3872} % *** Enter the CVPR Paper ID here

\ifcvprfinal\fi
\begin{document}

%%%%%%%%% TITLE
\title{Fine-grained Angular Contrastive Learning with Coarse Labels}

\author{
    Guy Bukchin$^{3,2}$,
    Eli Schwartz$^{1,2}$,
    Kate Saenko$^{1,4}$,\\
    Ori Shahar$^{3}$,
    Rogerio Feris$^{1}$,
    Raja Giryes*$^{2}$,
    Leonid Karlinsky\thanks{Equal contribution}\hspace{-1pt}*$^{1}$\\
    {\tt\small IBM Research AI$^{1}$, Tel-Aviv University$^{2}$, Penta-AI$^{3}$, Boston University$^{4}$}
}

% \and

\maketitle
%\thispagestyle{empty}

% \twocolumn[{%
% \renewcommand\twocolumn[1][]{#1}%
% \maketitle
% \begin{center}
%     \centering
%     \includegraphics[width=\textwidth]{figures/intro.pdf}
%     \captionof{figure}{(a) \emph{The \ourstaskfull (\ourstaskspace):}  During training we observe \textit{only coarse} class labels (in red, e.g. animals), while at test time we are expected to adapt our model to support the \textit{fine} classes using one or few samples. The fine classes may be sub-classes of the train classes (seen) or sub-classes of classes unseen during training; (b) our \oursspace method jointly employs coarse-supervised and intra-class self-supervised contrastive losses that would pull the model to different directions without our proposed \textit{Angular normalization} component that shifts the forces applied by the two losses to different planes and induces synergy between them leading to significant performance gains; (c) coarse-supervised baseline coarse classes tSNE; (d) coarse-supervised baseline fine sub-classes tSNE (arbitrary coarse class sub-classes), star marks the coarse class lin. classifier weight vector embedding, black arrows point to fine sub-classes centroids; (e) \oursspace coarse classes tSNE; (f) \oursspace fine sub-classes tSNE (same coarse class)}\label{fig:intro}
% \end{center}
% }]

%%%%%%%%% ABSTRACT
\begin{abstract}
Few-shot learning methods offer pre-training techniques optimized for easier later adaptation of the model to new classes (unseen during training) using one or a few examples.
This adaptivity to unseen classes is especially important for many practical applications where the pre-trained label space cannot remain fixed for effective use and the model needs to be "specialized" to support new categories on the fly. One particularly interesting scenario, essentially overlooked by the few-shot literature, is \ourstaskfull (\ourstask), where the training classes (e.g. animals) are of much `coarser granularity' than the target (test) classes (e.g. breeds). A very practical example of \ourstaskspace is when the target classes are sub-classes of the training classes. Intuitively, it is especially challenging as (both regular and few-shot) supervised pre-training tends to learn to ignore intra-class variability which is essential for separating sub-classes. In this paper, we introduce a novel 'Angular normalization' module that allows to effectively combine supervised and self-supervised contrastive pre-training to approach the proposed \ourstaskspace task, demonstrating significant gains in a broad study over multiple baselines and datasets. We hope that this work will help to pave the way for future research on this new, challenging, and very practical topic of \ourstaskspace classification.
\end{abstract}

%%%%%%%%% BODY TEXT
% \begin{figure}[H]
% \includegraphics[width=\linewidth]{figures/intro.pdf}
%     \captionof{figure}{(a) \emph{The \ourstaskfull (\ourstaskspace):}  During training we observe \textit{only coarse} class labels (in red, e.g. animals), while at test time we are expected to adapt our model to support the \textit{fine} classes using one or few samples. The fine classes may be sub-classes of the train classes (seen) or sub-classes of classes unseen during training; (b) our \oursspace method jointly employs coarse-supervised and intra-class self-supervised contrastive losses that would pull the model to different directions without our proposed \textit{Angular normalization} component that shifts the forces applied by the two losses to different planes and induces synergy between them leading to significant performance gains; (c) coarse-supervised baseline coarse classes tSNE; (d) coarse-supervised baseline fine sub-classes tSNE (arbitrary coarse class sub-classes), star marks the coarse class lin. classifier weight vector embedding, black arrows point to fine sub-classes centroids; (e) \oursspace coarse classes tSNE; (f) \oursspace fine sub-classes tSNE (same coarse class)}\label{fig:intro}
% \end{figure}
%
\begin{figure}
% \centering
\hspace{-18pt}
\includegraphics[width=1.15\linewidth]{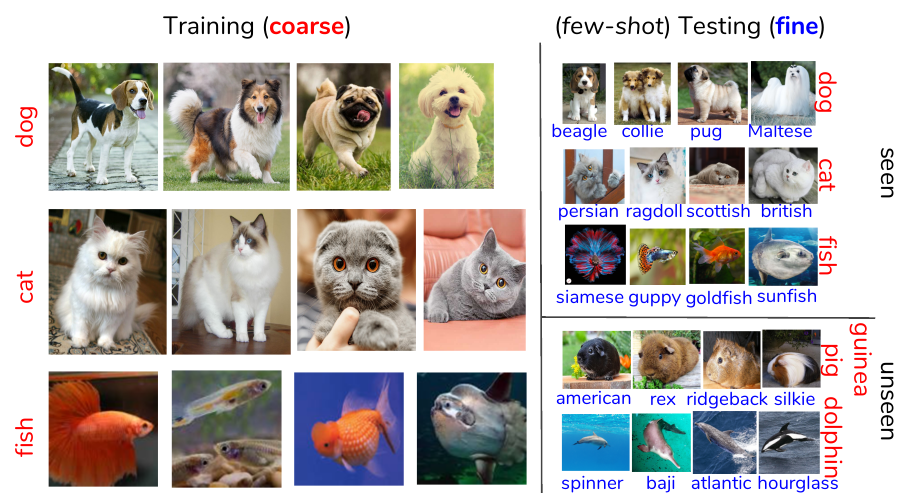}
\caption{\textbf{The \ourstaskfull (\ourstask):}  During training we observe \textit{only coarse} class labels (in red, e.g. animals), while at test time we are expected to adapt our model to support the \textit{fine} classes (in blue, e.g. breeds) using one or few samples. The fine classes may be sub-classes of the train classes (seen) or sub-classes of classes unseen during training.}
\label{fig:task}
\vspace{-15pt}
\end{figure}
%
%
% \begin{figure*}[h!]
%     \begin{center}
%     \includegraphics[width=\textwidth]{figures/intro.pdf}
%     \end{center}
%     \captionof{figure}{(a) Illustration of \ourstaskspace task, during training we observe \textit{only coarse} class labels (in red, e.g. animals), while at test time we are expected to adapt our model to support the \textit{fine} classes using one or few samples. The fine classes may be sub-classes of the train classes (seen) or sub-classes of classes unseen during training; (b) our \oursspace method jointly employs coarse-supervised and intra-class self-supervised contrastive losses that would pull the model to different directions without our proposed \textit{Angular normalization} component that shifts the forces applied by the two losses to different planes and induces synergy between them leading to significant performance gains; (c) coarse-supervised baseline coarse classes tSNE; (d) coarse-supervised baseline fine sub-classes tSNE (arbitrary coarse class sub-classes), star marks the coarse class lin. classifier weight vector embedding, black arrows point to fine sub-classes centroids; (e) \oursspace coarse classes tSNE; (f) \oursspace fine sub-classes tSNE (same coarse class)}\label{fig:intro}
% \end{figure*}
%
\vspace{-0.2cm}
\section{Introduction}\label{sec:intro}
\vspace{-0.1cm}
In the most commonly encountered learning scenario, supervised learning, a set of target (class) labels is provided for a set of samples (images) using which we train a model (CNN \cite{Krizhevsky2012,He2015,Tan2019} or Transformer \cite{HengshuangZhao2020, Ramachandran2019}) that casts these samples into some representation space from which predictions are made, e.g. using a linear classifier. Nevertheless, while supervised learning is a very common setting, in many practical applications the set of the target labels of interest is not static, and may change over time. One good example is few-shot learning \cite{Vinyals2016,Snell2017,Tian2020}, where a model is pre-trained in such a way that more classes could be added later with only very few additional labeled examples used for adapting the model to support these new classes.

However, in previous few-shot learning works most (if not all) of the new classes: (i) are separate from the classes the model already knows, in the sense that they either belong to a different branch of the class hierarchy or are siblings to the known classes; and (ii) are of same or similar level of granularity (same level of class hierarchy).
% This usually means that they share some attributes with some subsets of the known classes and through these shared attributes, hopefully retained by the feature space of the model, these new classes can be learned to be recognized (even from few samples) \cite{Tian2020}. 
But what about the very practical situation when the new classes are fine-grained sub-classes strictly included inside the known (coarse) classes being their descendants in the class taxonomy?
This situation typically occurs during the lifespan of the model when
the application requires separating some sub-classes of the current classes into separate classes and yet when the training dataset was created these (unknown in advance) sub-classes were not annotated. For example, this could occur in product specialization for product search, or during personalizing a generic model to a specific customer. Naturally, going back to re-labeling each time this occurs is much too costly to be an option.

\begin{figure}[t]
\centering
\includegraphics[width=1\linewidth]{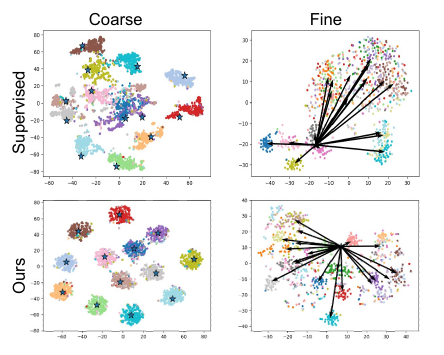}
\vspace{-20pt}
\caption{\textbf{Learned embedding tSNE visualization:} Top - coarse-supervised baseline, Bottom - ours (\ours). Left - coarse classes, right - fine sub-classes of one arbitrary coarse class. Stars are embeddings of the linear classifier (class) weight vectors, black arrows point from the class weight to the fine sub-classes centroids. Clearly, \oursspace induces order on the sub-classes arranging them nicely around the class weight and making them separable.}
\vspace{-15pt}
\label{fig:tsne}
\end{figure}
In this paper, we target the \ourstaskfull (\ourstask) task (Fig.~\ref{fig:task}) where a model pre-trained on a set of base classes (denoted as the `coarse' classes), needs to adapt on the fly to an additional set of target (`fine') classes of much `finer granularity' than the training classes. The target classes could be sub-classes of the base classes (a particularly interesting case), or they could be a separate set, yet requiring much stronger (than base classes) attention to fine-grained details in order to visually separate.
To be efficient, we want this adaptation to occur using only one or few samples of the fine (sub-)classes. Intuitively, this setup is particularly challenging for models pre-trained on the coarse classes in `the standard' supervised manner, as: \textbf{(a)} standard supervised learning losses do not care about the intra-class arrangement of the samples belonging to the same class in the model's feature space $\mathcal{F}$, as long as these samples are close to each other and the regions associated with different classes are separable (Fig. \ref{fig:tsne} top-left) - potentially causing the sub-classes to spread arbitrarily inside same-class-associated regions of $\mathcal{F}$ thus hindering their separability (Fig. \ref{fig:tsne} top-right); and \textbf{(b)} $\mathcal{F}$ is retaining the information on the attributes needed to predict the set of target `coarse' labels, while at the same time reducing intra-class variance and suppressing attributes not relevant to the task for better generalization, which may eliminate the intra-class distinctions between sub-classes (Fig. \ref{fig:tsne} top-right).

\begin{figure}[t]
\centering
\includegraphics[width=0.7\linewidth]{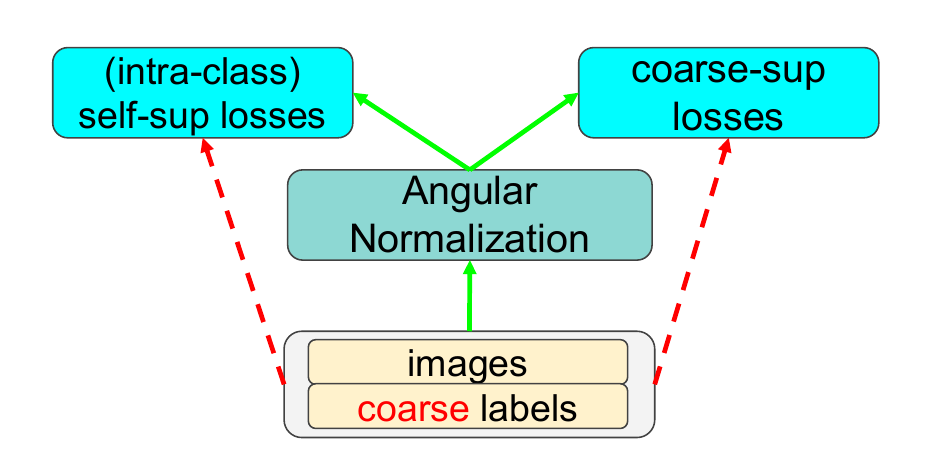}
\vspace{-7pt}
\caption{\textbf{\oursfull (\ours):} our method jointly employs inter-class supervised and intra-class self-supervised contrastive losses that would pull to different directions without our proposed \textit{Angular normalization} component that separates the forces applied by the two losses to different planes leading to significant performance gains.}
\vspace{-15pt}
\label{fig:concept}
\end{figure}
In contrast to supervised learning, recently emerged contrastive self-supervised methods \cite{chen2020simple,he2019moco,Caron2020,Grill2020} were proven highly instrumental in learning good features without any labels. These methods are able to pre-train effectively attaining almost the same representation (feature) quality as fully supervised counterparts, and even surpassing it when transferring to other tasks (e.g. detection \cite{he2019moco}). Even more importantly, these methods are optimizing features for 'instance recognition', retaining the information for identifying the fine details that separate instances between and within classes in the dataset, and thus likely also retaining features needed for effective sub-class recognition. That being said, contrastive methods have so far been mostly evaluated examining their ability for inter-class separation in a relatively favorable condition of an abundance of unlabeled data (e.g. ImageNet).
% So far, to the best of our knowledge, these have not been examined in fine-grained situations, with relatively few (unlabeled) \kate{this is confusing, are the 'few' samples unlabeled?} samples per sub-class of interest.
And yet, naive use of these methods for the \ourstaskspace task is sub-optimal. On their own, they lack the use of coarse labels supervision. And when naively used jointly with coarse-supervised losses, their lack of synergy with those losses leads to lower gains (Sec.~\ref{sec:ablation}).
%, as explained in more detail in Section \ref{sec:related} and verified in our experiments.

Building upon advances in contrastive self-supervised learning, we propose the \oursfull (\ours) approach for the \ourstaskspace task. It enables few-shot adaptation to fine-grained (sub-)classes using few examples, while pre-training using only coarse class labels. Our approach (Fig.~\ref{fig:concept}) effectively combines, in a multi-task manner, the supervised pre-training on the coarse classes that ensures inter-class separation, with contrastive self-supervised intra-class learning that facilitates the self-organization and separability of the fine sub-classes in the resulting feature space (Fig. \ref{fig:tsne} bottom). Our method features a novel angular normalization component that enhances the synergy between the supervised and contrastive self-supervised tasks, minimizing friction between them by separating their forces to different planes. We compare \oursspace to a diverse set of baselines and ablations, on multiple datasets,
% (BREEDS \cite{SanturkarS2020}, \textit{tiered}ImageNet \cite{Ren2018}, and CIFAR-100 \cite{Krizhevsky2009}), 
both underlining its effectiveness and providing a strong basis for future studies of the proposed \ourstaskspace task.

To summarize, our contribution is threefold: 
(i) we propose 
% (concurrently with another work)
the \ourstaskfull (\ourstask) task of training using only coarse class labels and adapting to support finer (sub-)classes with few (even one) examples; (ii) we propose the \oursspace approach for \ourstaskspace task, based on effective multi-task combination of supervised inter-class and self-supervised intra-class learning, featuring a novel angular normalization component to minimize friction and maximize the synergy between the two tasks; (iii) we offer extensive evaluation and analysis showing the strength of our proposed \oursspace approach on a variety of datasets and compared to a diverse set of baselines.% including the most prominent supervised and self-supervised methods.
\vspace{-0.1cm}
\section{Related Work}\label{sec:related}
\vspace{-0.1cm}
\textbf{Self-supervised learning.}
While the onset of deep-learning was pre-dominantly ruled by supervised learning  \cite{Krizhevsky2012,He2015,Tan2019}, recently many self-supervised representation learning methods have emerged. These works generate different self-induced (pretext) pseudo-labels for unlabeled data and drive the visual feature learning without any external supervision. Earlier works used predicting patch position \cite{Doersch_2015_ICCV}, image colorization \cite{zhang2016colorful}, jigsaw puzzles \cite{noroozi2016unsupervised}, image in-painting \cite{pathak2016context}, predicting image rotations \cite{gidaris2018unsupervised}, and others as pretext tasks. Yet, more recently, \cite{oord2018representation, tian2019contrastive, chen2020simple, he2019moco, chen2020mocov2, Grill2020, Caron2020} have demonstrated the power of  contrastive instance discrimination, significantly surpassing previous results and narrowing the gap with supervised methods.
SimCLR \cite{chen2020simple} defined positive pairs as two augmentations of the same image and contrasted them with other images of the same batch. Instead, MoCo \cite{he2019moco,chen2020mocov2} contrasted with samples extracted from a dynamic queue produced by a slowly progressing momentum encoder. 
% In the follow-up work, MoCoV2 \cite{chen2020mocov2} adopted some ideas from SimCLR to further improve MoCo performance. 
SWAV \cite{Caron2020} uses a clustering objective for computing the contrastive loss, BYOL \cite{Grill2020} replaces the contrastive InfoNCE loss with direct regression between positive pairs, essentially removing the need for negative samples, and \cite{XiaoT2020} explores the effect of different contrastive augmentation strategies. Interestingly, contrastive methods have recently shown promising results for domain adaptation \cite{Kang2019} and supervised learning \cite{Khosla2020}.
Intuitively, for solving our \ourstaskspace task both inter-class (between the coarse classes) and intra-class (within the classes) separation are jointly required. Supervised methods are better at inter-class separation, but are worse in intra-class separation (Fig. \ref{fig:tsne} top right), while contrastive self-supervised methods are better on intra-class and are worse on inter-class discrimination (Tab. \ref{tab:coarse_vs_fine}). In this paper, we show how to properly combine the two to enjoy the benefits of both.
% , where applying contrastive learning independently (per-class) would be problematic as each individual class data is usually a relatively small part of the whole dataset.
% Additionally, as both inter-class and intra-class separation are jointly required to solve the \ourstaskspace task, 
% Therefore,
% intuitively an interplay between coarse-supervised (e.g. cross entropy) and intra-class contrastive self-supervised losses is necessary. However, the supervised loss tries to pull the same-class instances closer together, while at the same time the intra-class contrastive loss pushes the instances of the same class apart, causing naive combinations of these losses to fail - a flaw addressed by the Angular normalization module proposed in our approach.

\textbf{Few-shot learning.} Meta-learning methods, which are very popular in the few-shot literature  \cite{Vinyals2016,Snell2017,relationnet,Li2019,Finn2017,Li2017,Zhou2018,Ravi2017,Munkhdalai2017,leo,closer_look,Schwartz2018,Oreshkin2018,Zhang2019,Zhang2019a,Dvornik2019,Lee2019,Hao2019,Zhang2020a,hou2019a,lifchitz2019a,metadapt}, learn from few-shot tasks (or episodes) rather than from individual labeled samples. Such tasks are small datasets, with a few labeled training (support) examples, and a few test (query) examples. The goal is to learn a model that at test time can be adapted to support novel categories, unseen during training on the base categories (with abundant train data). 
%In \cite{Gidaris2019} supervised with self-supervised pre-training are combined in order to boost the few-shot performance. 
In \cite{Qiao2019,Li2019_LST,Kim2019,Gidaris2019,Lichtenstein2020} additional unlabeled data is used, \cite{am3,Schwartz2019} leverage additional semantic information available for the classes, 
and \cite{Gidaris2019,KyleHsu2019,AntreasAntoniou2019,JongChyiSu2020} examine the usage of unsupervised or self-supervised training in the context of a standard few-shot learning. 
% and \cite{Gidaris2019,
% KyleHsu2019, % using unsupervised (clustering) and not sel-sup
% AntreasAntoniou2019,
% % Khodadadeh2019,
% JongChyiSu2020}
% examine the usage of self-supervision in the context of standard (fine) few-shot model pre-training. 
Recently, several works have noted that standard supervised pre-training on the base classes followed by simple fine-tuning attains (if done right) comparable and mostly better performance than the leading meta-learning methods \cite{Wang2019b,Tian2020,Lichtenstein2020}, even more strikingly so when the target (test) classes are in a different visual domain \cite{Guo2020}. Here, we build upon this intuition and do not use meta-learning for pre-training. Note though that in all these approaches, the base categories of the training set and the set of test categories are assumed to be of similar granularity (e.g. some ImageNet categories as the base and others as the target, or species of birds as the base and as the target, etc.). In particular, no method was proposed to tackle a commonly occurring (and hence very practical) situation of target classes being the \textit{sub-classes} of the base classes. Generally, situations when the target classes are from lower level of the classes hierarchy than the base classes have not been considered in the above works.

\begin{figure*}[t]
\begin{center}
\vspace{-10pt}
   \includegraphics[width=0.9\linewidth]{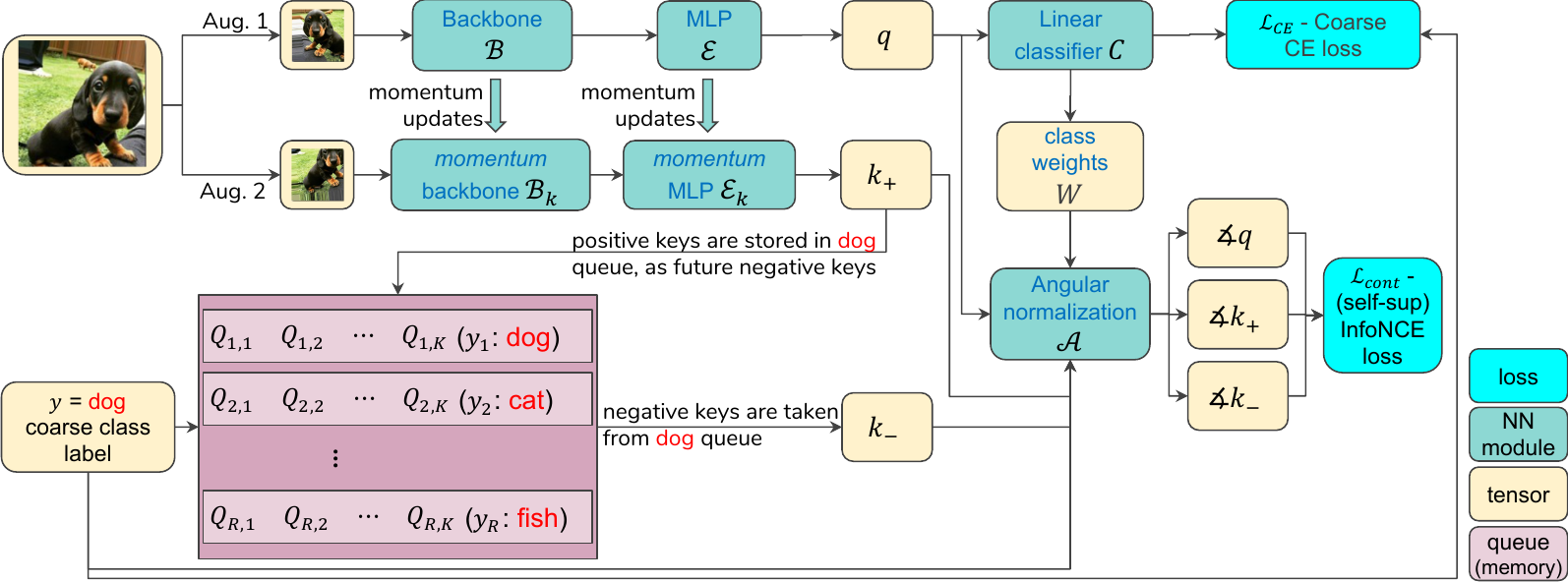}
\end{center}
\vspace{-10pt}
   \caption{\textbf{\oursspace method overview - training flow.} For illustrative purposes, showing a single image flow instead of a batch. The query $q$ and the positive key $k_{+}$ are computed from two random augmentations of the input image via the model ($\mathcal{B} \to \mathcal{E}$) and its momentum-updated counterpart ($\mathcal{B}_k \to \mathcal{E}_k$) \cite{he2019moco}. The negative keys $k_{-}$ are taken from the queue $\mathcal{Q}_y$ corresponding to the coarse class $y$ of the image. $q$ is classified to the coarse classes by linear classifier $C$, which is followed by the (supervised) CE loss $\mathcal{L}_{CE}$. The (self-supervised) InfoNCE loss $\mathcal{L}_{cont}$ is \textbf{not} applied directly on $q$, $k_{+}$, and $k_{-}$. Doing so would result in disagreement between supervised and self-supervised losses objectives. Instead, $q$, $k_{+}$, and $k_{-}$ are first normalized with the proposed \textit{Angular normalization} component $\mathcal{A}$ using the linear classifier $C$ weights $W$ corresponding to $y$. This disentangles the forces applied by the two losses, effectively leading to greater synergy between them and consequently to increased performance, as demonstrated by our experiments and ablations.}
\vspace{-15pt}
\label{fig:overview}
\end{figure*}
\textbf{Coarse and fine learning.} Relatively few works have considered learning problems entailing mixing of coarse and fine labels. Several works \cite{Ristin2015,YanmingGuo2018,Taherkhani2019} consider the partially fine-supervised training setting, where during training a mix of (equal \#) coarse- and fine- labeled samples is used for training. In contrast, in this work we focus on training using only coarse-labeled data, while fine categories are added at test time and from very few  examples (usually one). \cite{Hsieh2019} extends the partially fine-supervised setting to unbalanced splits between coarse- and fine- labeled samples via a MAML-like \cite{Finn2017} optimization, albeit in multi-label classification setting (and non-standard performance measure). Similarly, \cite{Robinson2020} also explores the unbalanced partially fine-supervised setting. Prototype propagation between coarse and fine classes on a known (in advance) class ontology graph is explored in \cite{LuLiu2019} in the few-shot context and in partially fine-supervised training setting, also assuming full knowledge of the classes graph (which also includes the test classes). Finally, a concurrent work of \cite{YangCoarse-to-FineClassification} focuses on a scenario similar to \ourstask. 
In \cite{YangCoarse-to-FineClassification} the learning is split into three separate consecutive steps: (i) feature learning using a combination of supervised learning with coarse labels and general batch-contrastive learning disregarding classes; (ii) greedy clustering of each coarse class into a set of (fine) pseudo-sub-classes in the resulting feature space; (iii) applying a meta-learning techniques to fine-tune the model to the set of generated pseudo-sub-classes. In contrast, in our \oursspace approach the model is trained end-to-end, contrastive learning is done within the coarse classes, and we propose a special angular loss component for significantly enhancing the supervised and self-supervised contrastive learning synergies. As a result, in section \ref{sec:unseen_coarse}, we obtain good gains over \cite{YangCoarse-to-FineClassification} on the same \textit{tiered}ImageNet test.
\vspace{-0.1cm}
\section{Method}\label{sec:method}
\vspace{-0.1cm}
\subsection{\ourstaskfull (\ourstask) task}\label{sec:task}
\vspace{-0.1cm}
Denote by $\mathcal{Y}_{coarse}=\{y_1,...,y_R\}$ a set of $R$ coarse training classes (e.g. kinds of animals: dog, cat, fish, ...), and let ${\mathcal{S}_{train}^{coarse}=\{(I_j, y_j) | y_j \in \mathcal{Y}_{coarse} \}_{j=1}^N}$ be a set of $N$ training images annotated (only) with $\mathcal{Y}_{coarse}$. Let ${\mathcal{Y}_{fine}=\{y_{1,1},...,y_{1,k_1} , y_{2,1},...,y_{2,k_2},...,y_{R,1},...y_{R,k_R}\}}$ be a set of fine sub-classes (e.g. animal breeds) of the coarse classes $\mathcal{Y}_{coarse}$. In our experiments we also explore the case when fine classes are sub-classes of unseen coarse classes. Let $\mathcal{B}$ be an encoder (CNN backbone) mapping images to a $d$-dimensional feature space $\mathcal{F} \subset \mathbb{R}^d$ (i.e. $\mathcal{B}(I_j)=F_j \in \mathcal{F}$) trained on $\mathcal{S}_{train}$. Provided at test time with a small $k$-shot training set for a subset $\mathcal{Y}_{fine}^m \subseteq \mathcal{Y}_{fine}$ of $m$ fine classes: $\mathcal{S}_{train}^{fine}=\{(I_r, y_r) | y_r \in  \mathcal{Y}_{fine}^m \}_{r=1}^{k \cdot m}$ our goal is to train a classifier $C : \mathcal{F} \to \mathcal{Y}_{fine}^m$ with maximal accuracy on the $\mathcal{Y}_{fine}^m$ fine classes test set. For example, $m$ could also be $\sum k_i$ making $\mathcal{Y}_{fine}^m=\mathcal{Y}_{fine}$ ('all-way'). Note that during training of $\mathcal{B}$ the set of fine sub-classes $\mathcal{Y}_{fine}$ is unknown. Also note that according to \cite{Wang2019b,Tian2020,Lichtenstein2020}, SOTA few-shot performance can be achieved even without modifying $\mathcal{B}$ when adapting to (unseen) test classes.

\vspace{-0.1cm}
\subsection{The (\ours) approach}
\vspace{-0.1cm}
At its core, our method focuses on learning $\mathcal{B}$ combining (with added synergy) supervised learning for inter-class separation of the coarse classes $\mathcal{Y}_{coarse}$ and contrastive self-supervised learning for separating the fine sub-classes within each coarse class ($\mathcal{Y}_{fine}$). The training architecture of \oursspace is illustrated in Fig. \ref{fig:overview}. Our model is comprised of: (i) a CNN encoder $\mathcal{B}:I \to \mathbb{R}^d$ with Global Average Pooling (GAP) on top (e.g. ResNet50 mapping images to $2048$-dimensional vectors); (ii) an MLP embedder module $\mathcal{E}: \mathbb{R}^d \to \mathbb{R}^e$ with $e < d$ (e.g. $2048 \to 2048 \to 128$), $\mathcal{E}$ also includes $L_2$ normalization of the final vector; (iii) second pair of (momentum) encoder $\mathcal{B}_k$ and (momentum) embedder $\mathcal{E}_k$ for encoding the positive keys in the contrastive objective that are momentum updated from $\mathcal{B}$ and $\mathcal{E}$ respectively (following \cite{he2019moco}); (iv) a linear classifier $C: \mathbb{R}^e \to \mathcal{Y}_{coarse}$ without bias and $W \in \mathbb{R}^{R \times e}$ its weight matrix (so $W \cdot \mathcal{E}(\mathcal{B}(I))$ are the $R$ coarse classes logits of $C$); (v) a set of per-class 'negative-instance' queues $\{\mathcal{Q}_i\}_{i=1}^R$, with each queue: $\mathcal{Q}_i\in\mathbb{R}^{e\times K}$ of length $K$ (different from \cite{he2019moco} that utilized a single queue for the entire dataset); and (vi) an \textit{Angular normalization} module explained below, that is used for inducing synergy between the supervised and self-supervised contrastive losses by disentangling the forces they apply on samples in the feature space.

The training (Fig. \ref{fig:overview}) proceeds in batches, but for clarity here we describe the training process for a single training image $I$ annotated with a coarse class label $y \in \mathcal{Y}_{coarse}$. Abusing notation let $y$ also denote the index of the coarse class in $\mathcal{Y}_{coarse}$. We first augment $I$ twice to get $I_q$ and $I_k$ which are then passed through the corresponding encoders and embedders to retrieve  $q=\mathcal{E}(\mathcal{B}(I_q))$ and $k_{+}=\mathcal{E}_k(\mathcal{B}_k(I_k))$. We also set $k_{-}=\mathcal{Q}_y$ (being 'momentum' embeddings of previously encountered samples of the same coarse class $y$, at the end of the training cycle $k_{+}$ is also added to $\mathcal{Q}_y$). Our two loss functions are: $\mathcal{L}_{CE}(C(q),y)$ being the coarse-supervised softmax Cross Entropy (CE) on $C(q)=W \cdot q$ logits and $1$-hot vector for $y$ coarse class label; and $\mathcal{L}_{cont}(q,k_{+},k_{-})$ being the $y$ class-specific self-supervised contrastive InfoNCE loss applied to query $q$ positive key $k_{+}$ and negative keys $k_{-}$. However, if used naively, $\mathcal{L}_{CE}$ would try to push $q$ towards the other samples of the same class $y$, while at the same time $\mathcal{L}_{cont}$ would try to push it away from them (as $k_{-}$ represents the other samples of the class). This would diminish the synergy between the losses and, as shown in the ablation study \ref{sec:ablation}, would result in significant performance drop on the the \ourstaskspace task.

\begin{figure}
\vspace{-10pt}
\centering
\includegraphics[width=0.8\linewidth]{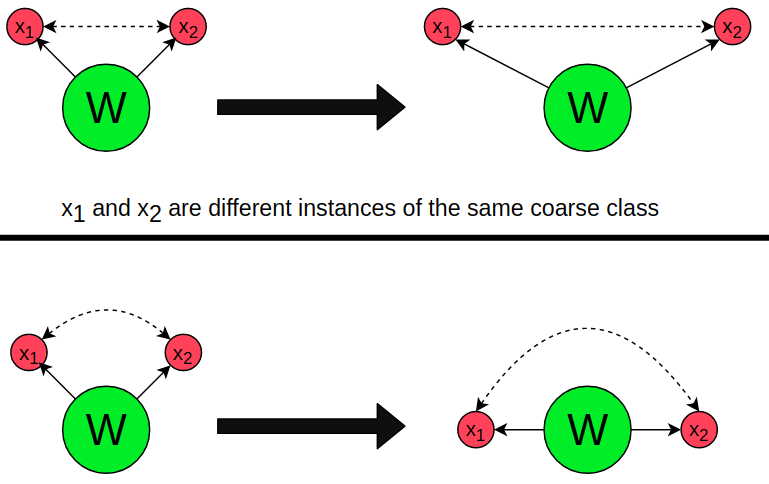}
% \vspace{-10pt}
\caption{\emph{The angular normalization effect.} \textbf{Top:} without normalization minimizing the contrastive loss $\mathcal{L}_{cont}$ pushes the samples of the same coarse class $x_1$, $x_2$ (red) away from each other and thus away from the class weight vector $W$ (green) increasing the supervised loss $\mathcal{L}_{CE}$;
\textbf{Bottom:} after angular normalization, $\mathcal{L}_{cont}$ operates on angles in the orbit around the class weight, not affecting the distance between the weight and the samples.}
\vspace{-15pt}
\label{fig:angular}
\end{figure}
{\bf Angular normalization.} To improve this synergy, we propose a new module which we name 'angular normalization'. For a given image $I$ with embedding $q$ and coarse label $y$, the logit for class $y$ in classifier $C$ is $W_y \cdot q$, where $W_y$ is the $y^{th}$ row of $W$. Thus, the supervised loss $\mathcal{L}_{CE}(C(q),y)$ is minimized when $W_y \cdot q$ is maximized and $W_{i \ne y} \cdot q$ are minimized, or in other words, when $q$ (unit vector, as the embedder $\mathcal{E}$ ends with $L_2$ normalization) shifts towards being in the direction of $W_y$. And this is the same for all images of class $y$ essentially encouraging their collapse to $W_y/||W_y||$ (the unit vector closest to $W_y$). But this collapse is in direct conflict of interests with the $y$ class-specific InfoNCE contrastive loss $\mathcal{L}_{cont}(q,k_{+},k_{-})$ that tries to push $y$'s samples away from each other (Fig. \ref{fig:angular} top). To solve this, we propose a simple method that can be used to induce synergy between $\mathcal{L}_{CE}$ and $\mathcal{L}_{cont}$. We define the $y$-class specific angular normalizaton: %$\mathcal{A}(\cdot,W,y)$:
\begin{equation}
\mathcal{A}(x,W,y)=\angle x=\frac{\frac{x}{\rVert{x}\rVert} - \frac{\mathcal{W}_y}{\rVert\mathcal{W}_y\rVert}}{\rVert\frac{x}{\rVert{x}\rVert} - \frac{\mathcal{W}_y}{\rVert\mathcal{W}_y\rVert}\rVert}.
\end{equation}
which converts any unit vector $x/||x||$ into a unit vector representing its angle around $W_y/||W_y||$. With the above definition of angular normalization $\mathcal{A}$, we replace the $q$, $k_{+}$, and $k_{-}$ in $\mathcal{L}_{cont}$ with their $y$-class-specific normalized versions:
%Using the angular normalization component $\mathcal{A}$ which is used to produce:
\begin{align}
    \angle q &= \mathcal{A}(q,W,y) \\
    \angle k_{+} &= \mathcal{A}(k_{+},W,y) \\
    \angle k_{-} &= \mathcal{A}(k_{-},W,y)
\end{align}
%applying $\mathcal{L}_{cont}$ to which will shift the contrastive forces operating on the sample to a different plane not interfering with the forces applied by $\mathcal{L}_{CE}$.
and obtain our \textbf{final (full) loss function}: 
\begin{equation}
\mathcal{L}=\mathcal{L}_{CE}(C(q),y) + \mathcal{L}_{cont}(\angle q, \angle k_{+}, \angle k_{-})
\end{equation}
where the angular normalized contrastive loss $\mathcal{L}_{cont}(\angle q, \angle k_{+}, \angle k_{-})$ operates in the space of angles in the 'orbit' around $W_y/||W_y||$, thus not interfering with the 'drive to collapse to $W_y/||W_y||$' dictated by the $\mathcal{L}_{CE}$ loss (Fig. \ref{fig:angular} bottom). An additional intuitive benefit of the $\mathcal{A}$ normalization is that it ignores the distance to the (normalized) class weight vector, thus protecting $\mathcal{L}_{cont}$ from bias caused by different 'tighter' or 'looser' sub-classes.

\subsection{Few-shot testing on fine classes}\label{Few-shot testing on fine classes}
At test time, only the encoder $\mathcal{B}$ followed by $L_2$ normalization is retained as the feature extractor and following \cite{ChenX2020MoCoV2} the MLP embedder $\mathcal{E}$ is dropped (for higher performance). According to our definition of \ourstaskspace task, only a small $k$-shot and $m$-way training set $\mathcal{S}_{train}^{fine}$ is available for adapting the model to support the fine-classes. For the few-shot classifier we use the method of \cite{Tian2020}. For every few-shot episode, we create 5 augmented copies for every support sample, and train a logistic regression model on the support set encoded using $\mathcal{B}$ followed by $L_2$ normalization. The model with the resulting logistic regression classifier on top is then used to classify the query samples of the episode.

%this is indeed part of the evaluation protocol:
% We report results for \textbf{5-way 1-shot} and \textbf{all-way 1-shot} (15-query) tests. We follow \cite{Tian2020}: we run 1000 episodes and report the mean accuracy and the 95\% confidence interval.

% move to the place where we discuss results table for now
% All models are pretrained on a training set, and for evaluation their linear classification layer \textbf{is dropped}. This detail is trivial for coarse baselines, since the classes they see in training and testing are completely different. However, for fine baselines, which theoretically should keep the classification layer at test time, this is detrimental. We do this so that the analysis focuses on feature generalization using similar few-shot learning methods, as opposed to comparing a model that learns via few-shot learning with a model that learns a linear classifier that sees all of the data.
\vspace{-0.1cm}
\section{Experiments}\label{sec:results}
\vspace{-0.1cm}
\subsection{Datasets}\label{sec:datasets}
\vspace{-0.1cm}
Our experiments were performed on: (i) \textbf{BREEDS} \cite{SanturkarS2020}, four datasets derived from ImageNet with class hierarchy re-calibrated by \cite{SanturkarS2020} so classes on same hierarchy level are of the same visual granularity (not so in the WordNet hierarchy); (ii) \textbf{CIFAR-100} \cite{Krizhevsky2009}; and (iii) \textbf{tieredImageNet} \cite{Ren2018},
a subset of ImageNet, with train/val/test built from different coarse classes. Datasets are summarised in Table \ref{tab:breeds_info}.

\begin{table}[t]
    \centering
    \resizebox{1.0\linewidth}{!}{%
    \begin{tabular}{lcccccc}
    \toprule
         Dataset & L17 & NL26 & E13 & E30 & CIFAR100 & Tiered \\
    \midrule
         \# Coarse classes & 17 & 26 & 13 & 30 & 20 & 20/6 \\
         \# Fine classes & 68 & 104 & 260 & 240 & 100 & 351/160 \\
         \# Train images & 88K & 132K & 334K & 307K & 50K & 448K \\
         \# Test images & 3.4K & 5.2K & 13K & 12K & 10K & 206K \\
         Image resolution & 224 & 224 & 224 & 224 & 32 & 84 \\
    \bottomrule
    \end{tabular}}
    \caption{\textbf{Datasets:} L17, NL26, E13 \& E30 are the Living17, NonLiving17, Entity13 \& Entity30 from BREEDS. For Tiered the train/val/test classes are non overlapping; For CIFAR-100 and BREEDS val was 10\% of the train, Tiered has pre-defined val set.}
    \vspace{-15pt}
    \label{tab:breeds_info}
\end{table}

\begin{table*}[bth!]
\resizebox{\textwidth}{!}{%
\begin{tabular}{l@{\hskip 12pt}c@{\hskip 6pt}c@{\hskip 12pt}c@{\hskip 6pt}c@{\hskip 12pt}c@{\hskip 6pt}c@{\hskip 12pt}c@{\hskip 6pt}c}
\toprule
 &
  \multicolumn{2}{c}{LIVING-17} &
  \multicolumn{2}{c}{NONLIVING-26} &
  \multicolumn{2}{c}{ENTITY-13} &
  \multicolumn{2}{c}{ENTITY-30} \\
Method              & 5-way        & all-way      & 5-way        & all-way      & 5-way        & all-way      & 5-way        & all-way      \\
\midrule

Fine (upper-bound) & 91.10 \scriptsize{$\pm$ 0.47}  & 58.95 \scriptsize{$\pm$ 0.16} & 85.25 \scriptsize{$\pm$ 0.49} & 47.68 \scriptsize{$\pm$ 0.13} & 91.01 \scriptsize{$\pm$ 0.39} & 50.19 \scriptsize{$\pm$ 0.08} & 91.65 \scriptsize{$\pm$ 0.41} & 56.54 \scriptsize{$\pm$ 0.09} \\
Fine$+$ (upper-bound)    & 78.39 \scriptsize{$\pm$ 0.64} & 46.92 \scriptsize{$\pm$ 0.16} & 74.95 \scriptsize{$\pm$ 0.57} & 39.57 \scriptsize{$\pm$ 0.11} & 85.98 \scriptsize{$\pm$ 0.55} & 47.87 \scriptsize{$\pm$ 0.09} & 85.43 \scriptsize{$\pm$ 0.57} & 45.87 \scriptsize{$\pm$ 0.09} \\
\midrule
MoCoV2         & 56.66 \scriptsize{$\pm$ 0.70}  & 18.57 \scriptsize{$\pm$ 0.11} & 63.51 \scriptsize{$\pm$ 0.75} & 21.07 \scriptsize{$\pm$ 0.11} & 82.00 \scriptsize{$\pm$ 0.67}  & 33.06 \scriptsize{$\pm$ 0.07} & 80.37 \scriptsize{$\pm$ 0.62} & 28.62 \scriptsize{$\pm$ 0.06} \\
MoCoV2-ImageNet \cite{chen2020mocov2} & 82.21 \scriptsize{$\pm$ 0.73} & 40.29 \scriptsize{$\pm$ 0.14} & 77.07 \scriptsize{$\pm$ 0.78} & 34.78 \scriptsize{$\pm$ 0.13} & 85.24 $\pm$ 0.6  & 35.62 \scriptsize{$\pm$ 0.08} & 83.06 \scriptsize{$\pm$ 0.62} & 31.73 \scriptsize{$\pm$ 0.08} \\
SWAV-ImageNet \cite{Caron2020} & 79.83 \scriptsize{$\pm$ 0.65} & 38.79 \scriptsize{$\pm$ 0.15} & 76.26 \scriptsize{$\pm$ 0.71} & 33.94 \scriptsize{$\pm$ 0.11} & 81.15 \scriptsize{$\pm$ 0.65} & 33.57 \scriptsize{$\pm$ 0.07} & 79.91 \scriptsize{$\pm$ 0.54} & 31.15 \scriptsize{$\pm$ 0.07} \\
Coarse        & 85.12 \scriptsize{$\pm$ 0.74} & 33.83 \scriptsize{$\pm$ 0.10}  & 83.53 \scriptsize{$\pm$ 0.64} & 33.52 \scriptsize{$\pm$ 0.11} & 82.33 \scriptsize{$\pm$ 0.61} & 17.49 \scriptsize{$\pm$ 0.04} & 87.03 \scriptsize{$\pm$ 0.54} & 24.01 \scriptsize{$\pm$ 0.06} \\
Coarse$+$  & 79.29 \scriptsize{$\pm$ 0.65} & 37.44 \scriptsize{$\pm$ 0.12} & 75.91 \scriptsize{$\pm$ 0.66} & 36.80 \scriptsize{$\pm$ 0.11}  & 83.23 \scriptsize{$\pm$ 0.66} & 31.15 \scriptsize{$\pm$ 0.07} & 84.81 \scriptsize{$\pm$ 0.61} & 33.22 \scriptsize{$\pm$ 0.08} \\ \midrule
\oursspace (ours) &
  \textbf{89.23 \scriptsize{$\pm$ 0.55}} &
  \textbf{45.14 \scriptsize{$\pm$ 0.12}} &
  \textbf{86.23 \scriptsize{$\pm$ 0.54}} &
  \textbf{43.10 \scriptsize{$\pm$ 0.11}} &
  \textbf{90.58 \scriptsize{$\pm$ 0.54}} &
  \textbf{42.29 \scriptsize{$\pm$ 0.08}} &
  \textbf{88.12 \scriptsize{$\pm$ 0.54}} &
  \textbf{41.79 \scriptsize{$\pm$ 0.08}} \\ 
 \bottomrule
\end{tabular}}
\caption{Results for different baselines on the four BREEDS datasets. Top section contains models trained with fine-grained labels that serve as upper bounds, middle section contains baselines, and the bottom section is our results.}
    \vspace{-15pt}
\label{tab:breeds}
\end{table*}
%

% The \textbf{BREEDS} \cite{SanturkarS2020} are four datasets derived from ImageNet with class hierarchy re-calibrated (using crowd-sourcing) so classes on same hierarchy level are of the same visual granularity (not necessarily so in the original WordNet hierarchy). Table \ref{tab:breeds_info} summarizes BREEDS metadata.
% The datasets comprising BEEDS are: (i) \textbf{LIVING-17:} 88K train / 3.4K test samples, 17 coarse classes, 4 fine classes per coarse class; (ii) \textbf{NONLIVING-26:} 132K train / 5.2K test samples, 26 coarse classes, 4 fine classes per coarse class; (iii) \textbf{ENTITY-13:} 334K train / 13K test samples, 13 coarse classes, 20 fine classes per coarse class; (iv) \textbf{ENTITY-30:} 307K train / 12K test, 30 coarse classes, 8 fine classes per coarse class. 
% For all BREEDS datasets, 10\% of the data was used for validation.

% The \textbf{CIFAR-100} dataset \cite{Krizhevsky2009} has 20 coarse classes, each with 5 fine sub-classes. All images are of $32\times32$ resolution. It has 50K training images, of which 5K are used for validation, and 10K are test images.

% The \textbf{\textit{tiered}ImageNet} dataset \cite{Ren2018} is a subset of ImageNet, which contains 608 fine classes grouped into 34 coarse classes. These coarse classes are divided between the partitions: train (448K images, 351 fine, 20 coarse), validation (124K images, 97 fine, 6 coarse) and test (206K images, 160 fine, 8 coarse). No classes are shared between the partitions. Images are of $84\times84$ resolution.

\vspace{-0.1cm}
\subsection{Baselines}\label{sec:baselines}
\vspace{-0.1cm}
As \ourstaskspace task is new, we propose a diverse set of natural baselines and upper bounds for the task. We also compare to the concurrent work of \cite{YangCoarse-to-FineClassification} where applicable. For fair comparison, we use $200$ train epochs for all models (ours, baselines). The effect of longer training is explored in section \ref{sec:epochs}. Hyper-parameters of all compared methods were tuned on val sets. For the baselines and upper bounds we also used the best training practices from \cite{Tian2020}, including self-distillation, which consistently improved their performance. For fairness, in each experiment all compared methods use the same backbone architecture (for the encoder $\mathcal{B}$).

\textbf{Coarse Baselines:} models trained using \textbf{coarse labels} and with $\mathcal{L}_{CE}$ loss. We consider two such models: (i) 'Coarse' being the encoder $\mathcal{B}$ followed by a linear classifier $C$; (ii) 'Coarse$+$' being $\mathcal{B} \to \mathcal{E} \to C$ which has the same number of learned parameters as our \oursspace model. At test time, the MLP embedder $\mathcal{E}$ is dropped from Coarse$+$ for higher performance (same as for \oursspace).
% When few-shot testing, for both Coarse and Coarse$+$ baselines, we drop the linear classifier $C$, and for Coarse$+$ also drop the embedder MLP $\mathcal{E}$, same as we do for the \oursspace model. In all experiments, \textit{not dropping} $\mathcal{E}$ in Coarse$+$ lead to lower performance (not reported for space).

\textbf{Self-supervised Baselines:} 
% The MocoV2 \cite{chen2020mocov2} and SWAV \cite{Caron2020} SOTA contrastive self-supervised methods. As, similarly to MocoV2, \oursspace also includes momentum update and queues components, we compare to MocoV2 in two ways:
 Two baselines of MoCoV2 \cite{chen2020mocov2}: (i) 'MocoV2' is using the \cite{chen2020mocov2} official code to train on respective training sets; (ii) 'MocoV2-ImageNet' is the official full ImageNet pre-trained model of \cite{chen2020mocov2}. Similarly, 'SWAV-ImageNet' is the official model of \cite{Caron2020}. 
%  Both MocoV2-ImageNet and SWAV-ImageNet are compared to \oursspace on the ImageNet resolution datasets of BREEDS. 
 Note that full ImageNet pre-trained models saw more data during training than did \oursspace, and yet, interestingly, \oursspace attains better results. Finally, 'naive combination' of supervised and self-supervised losses without our angular normalization $\mathcal{A}$ is explored in ablation (Sec. \ref{sec:ablation}). Due to lack of space, additional baselines are provided in Supplementary.

\textbf{Fine Upper-Bound:} a natural performance upper-bounds for \oursspace are the $\mathcal{B} \to C$ and $\mathcal{B} \to \mathcal{E} \to C$ models trained on the fine labels of the respective training sets (hidden from \oursspace according to the \ourstaskspace task definition). To be consistent with the coarse baselines naming convention, we call them 'Fine' and 'Fine$+$' respectively. To match the evaluation setting, for both 'Fine' and 'Fine$+$' we also drop the classifier $C$ and embedder $\mathcal{E}$ when few-shot testing.
% (also essential in \textit{tiered}ImageNet experiments where test classes are different from train ones).

\vspace{-0.1cm}
\subsection{Additional implementation details}
\vspace{-0.1cm}
The encoder $\mathcal{B}$ was: ResNet-50 \cite{He2015} for the $224\times224$ datasets (BREEDS \cite{SanturkarS2020}), and ResNet-12 for the small resolution datasets (CIFAR-100 \cite{Krizhevsky2009} and \textit{tiered}ImageNet \cite{Ren2018}), as is common in the self-supervised and few-shot works respectively. The output dim of these networks is $d=2048$ or $d=640$, respectively. Our MLP embedder $\mathcal{E}$ consisted of two stacked fully connected layers with ReLu activation: $d \to d \to e$, we used $e=128$ in all experiments. We used cosine-annealing with warm restarts schedule \cite{Loshchilov2016} with 20 epochs per cycle. We trained on 4 V100 GPUs, with a batch size $b=256$ and base learning rate $lr=0.03$ for BREEDS, and $b=1024$ and $lr=0.12$ for CIFAR-100 and \textit{tiered}ImageNet. We used $wd=1e^{-4}$ weight decay. We used queue size $K=65536$, InfoNCE temperature of $\tau=0.2$, and $m=0.999$ for the momentum encoder ($m=0.99$ for CIFAR-100). 
% on BREEDS and \textit{tiered}ImageNet, and $m=0.99$ for CIFAR-100. 
All hyper-parameters were optimized on val sets, using the same optimization for our method and the baselines. Code: \textcolor{blue}{github.com/guybuk/ANCOR}.

\vspace{-0.1cm}
\subsection{Results}
\vspace{-0.1cm}
We report results for \textbf{5-way $k$-shot} and \textbf{all-way $k$-shot} $15$-query tests. Following \cite{Tian2020} we evaluate on 1000 random episodes and report the mean accuracy and the 95\% confidence interval. Unless explicitly stated $k=1$. Effect of more shots is evaluated in section \ref{sec:shots}. The test classes of each episode are a random subset of the set of fine classes $\mathcal{Y}_{fine}$ (or all of them in all-way tests). In section \ref{sec:where}, in order to further investigate the sources of our improvements, we evaluate a special case of 'intra-class' testing, when all the categories of the episode belong to the same (randomly sampled) coarse class.

\vspace{-10pt}
\subsubsection{Unseen fine sub-classes of seen coarse classes}\label{sec:seen_coarse}
\vspace{-0.1cm}
We first evaluate the core use-case of the \ourstaskspace task, namely training on coarse classes $\mathcal{Y}_{coarse}$ and generalizing to fine sub-classes of those classes $\mathcal{Y}_{fine}$ as defined in section \ref{sec:task}. We used BREEDS and CIFAR-100 for this evaluation, and the results are reported in Tables \ref{tab:breeds} and \ref{tab:cifar100} respectively. As can be seen, \oursspace significantly outperforms the coarse baselines across all datasets, in both the 5-way and the all-way tests (e.g. on BREEDS, by over 10\% on 5-way and over 5\% on all-way). Notably, in NONLIVING-26 5-way, our model surpasses even the Fine models. Moreover, on BREEDS we observe large gains (over 5\% in 5-way and over 6\% in all-way) over full ImageNet pre-trained self-supervised baselines in all datasets. This suggests that even significantly larger and more diverse training data available to those models is not sufficient to bridge the gap of coarse classes supervision which is needed for the \ourstask, and is effectively utilized by \oursspace in good synergy with the self-supervised objective (due to our angular component). In the supplementary we explore a 'sub-population shift' variant of this scenario where the  sub-classes appearing in training (with the coarse class label only) are non-overlapping with those appearing in the test.

% \begin{table}[]
% \begin{tabular}{|l|cc|}
% \hline
%              & \multicolumn{2}{c|}{CIFAR-100} \\
%              & 5-way          & all-way       \\ \hline\hline
% Fine         & 74.36 $\pm$ 0.68   & 28.82 $\pm$ 0.11  \\
% Fine (mlp)   & 69.65 $\pm$ 0.67   & 27.0 $\pm$ 0.11   \\\hdashline
% Coarse       & 74.4 $\pm$ 0.7     & 27.37 $\pm$ 0.11  \\
% Coarse (mlp) & 70.69 $\pm$ 0.69   & 26.16 $\pm$ 0.1   \\ \hline
% Ours         & \textbf{74.56 $\pm$ 0.7}    & \textbf{29.84 $\pm$ 0.11}  \\ \hline
% \end{tabular}
% \end{table}

\begin{table}[t]
\begin{center}
\resizebox{0.75\linewidth}{!}{%
\begin{tabular}{lcc}
\toprule
            %  & \multicolumn{2}{c}{CIFAR-100} \\
             & 5-way          & all-way       \\
\midrule
% Upper-bound: \\
Fine (upper-bound)       & 74.36 \scriptsize{$\pm$ 0.68}   & 28.82 \scriptsize{$\pm$ 0.11}  \\
Fine$+$ (upper-bound)  & 69.65 \scriptsize{$\pm$ 0.67}   & 27.00 \scriptsize{$\pm$ 0.11}   \\
\midrule
MoCo V2 & 48.07 \scriptsize{$\pm$ 0.68} & 10.61 \scriptsize{$\pm$ 0.06} \\
Coarse       & 74.40 \scriptsize{$\pm$ 0.70}    & 27.37 \scriptsize{$\pm$ 0.11}  \\
Coarse$+$ & 70.69 \scriptsize{$\pm$ 0.69}   & 26.16 \scriptsize{$\pm$ 0.10}   \\
\midrule
\oursspace (ours)         & \textbf{74.56 \scriptsize{$\pm$ 0.70}}    & \textbf{29.84 \scriptsize{$\pm$ 0.11}}  \\ 
\bottomrule
\end{tabular}}
\captionof{table}{Results on CIFAR-100.}
    \vspace{-25pt}
\label{tab:cifar100}
\end{center}
\end{table}

\vspace{-10pt}
\subsubsection{Unseen fine sub-classes of unseen coarse classes}\label{sec:unseen_coarse}
\vspace{-0.1cm}
We use the \textit{tiered}ImageNet dataset to evaluate the second use-case for \ourstaskspace task: when the fine classes $\mathcal{Y}_{fine}$ are not sub-classes of the training coarse classes $\mathcal{Y}_{coarse}$, and in fact belong to a different branch of the classes taxonomy, and yet $\mathcal{Y}_{fine}$ are of significantly higher visual granularity then $\mathcal{Y}_{coarse}$. We train using only the coarse labels of \textit{tiered}ImageNet train classes, and evaluate on its standard test set (with its fine labels). As \textit{tiered}ImageNet is a standard few-shot benchmark, here we compare to the few-shot SOTA methods \cite{Tian2020,hou2019a,Zhang2020a} trained using coarse train labels. The results of this experiment are summarized in Table \ref{tab:tiered} showing significant advantage of \oursspace over the baselines. Interestingly, the self-supervised MoCoV2 lags (significantly) behind, underlining the benfit of additional coarse supervision even in situation when test classes are descendants of different coarse classes. Notably, \oursspace also has a good 3\%-5\% advantage over the results of the concurrent work of \cite{YangCoarse-to-FineClassification} also dealing with coarse-and-fine few-shot interplay and performing the same experiment.

%
% \begin{table}[]
% \resizebox{1.0\linewidth}{!}{%
% \begin{tabular}{llll}
% \toprule
%                         %   & \multicolumn{3}{c}{\textit{tiered}ImageNet}                                 \\
%   & 5-way         & 5-way          & All-way        \\ 
%   & 1-shot         & 5-shot          & 1-shot        \\ 
%   \midrule
% % Upper-bound: \\
% Fine \scriptsize{(upper-bound)} \cite{Tian2020}  & 70.15 \scriptsize{$\pm$ 0.70}         & 84.96 \scriptsize{$\pm$ 0.47}          & 15.42 \scriptsize{$\pm$ 0.06}          \\ 
% % (upper-bound) \\
% \midrule
% Coarse & 61.61 \scriptsize{$\pm$ 0.74}         & 75.72 \scriptsize{$\pm$ 0.56}          & 8.42 \scriptsize{$\pm$ 0.04}           \\
% Coarse$+$ & - & - & -          \\
% MoCoV2 & 53.19 \scriptsize{$\pm$ 0.68} & 70.90 \scriptsize{$\pm$ 0.58} & 8.33 \scriptsize{$\pm$ 0.04} \\
% BDE-MetaBL \cite{YangCoarse-to-FineClassification}                & 60.54 \scriptsize{$\pm$ 0.79}         & 75.22 \scriptsize{$\pm$ 0.63}          & N/A                   \\ 
% \midrule
% \oursspace (ours)        & \textbf{63.54 \scriptsize{$\pm$ 0.70}} & \textbf{80.12 \scriptsize{$\pm$ 0.53}} & \textbf{11.97 \scriptsize{$\pm$ 0.06}} \\ 
% \bottomrule
% \end{tabular}}
% \caption{Results on \textit{tiered}ImageNet.}
%     \vspace{-10pt}
% \label{tab:tiered}
% \end{table}
%
\begin{table}[b]
\vspace{-15pt}
\resizebox{1.0\linewidth}{!}{%
\begin{tabular}{llll}
\toprule
                        %   & \multicolumn{3}{c}{\textit{tiered}ImageNet}                                 \\
   & 5-way         & 5-way          & All-way        \\ 
   & 1-shot         & 5-shot          & 1-shot        \\ 
   \midrule
% Upper-bound: \\
\cite{Tian2020} fine \scriptsize{(upper-bound)}  & 70.15 \scriptsize{$\pm$ 0.70}         & 84.96 \scriptsize{$\pm$ 0.47}          & 15.42 \scriptsize{$\pm$ 0.06}          \\ 
% (upper-bound) \\
\midrule
\cite{Tian2020} coarse & 61.61 \scriptsize{$\pm$ 0.74}         & 75.72 \scriptsize{$\pm$ 0.56}          & 8.42 \scriptsize{$\pm$ 0.04}           \\
MoCoV2 & 53.19 \scriptsize{$\pm$ 0.68} & 70.90 \scriptsize{$\pm$ 0.58} & 8.33 \scriptsize{$\pm$ 0.04} \\
CAN \cite{hou2019a}               & 56.91 \scriptsize{$\pm$ 0.55}         & 69.76 \scriptsize{$\pm$ 0.46}          &   6.29 \scriptsize{$\pm$ 0.08}                 \\ 
BDE-MetaBL \cite{YangCoarse-to-FineClassification}                & 60.54 \scriptsize{$\pm$ 0.79}         & 75.22 \scriptsize{$\pm$ 0.63}          & N/A                   \\ 
DeepEMD \cite{Zhang2020a}                & 62.84 \scriptsize{$\pm$ 0.71}         & 76.95 \scriptsize{$\pm$ 0.76}          &   9.65 \scriptsize{$\pm$ 0.15}                 \\ 
\midrule
\oursspace (ours)        & \textbf{63.54 \scriptsize{$\pm$ 0.70}} & \textbf{80.12 \scriptsize{$\pm$ 0.53}} & \textbf{11.97 \scriptsize{$\pm$ 0.06}} \\ 
\bottomrule
\end{tabular}}
\caption{Results on \textit{tiered}ImageNet.}
    \vspace{-10pt}
\label{tab:tiered}
\end{table}
\vspace{-0.1cm}
\subsection{Analysis}\label{sec:analysis}
\vspace{-0.1cm}
We used the LIVING-17 dataset for the analysis.
\vspace{-10pt}
\subsubsection{Ablation study}\label{sec:ablation}
\vspace{-0.1cm}
Here we evaluate several design choices of the \oursspace approach. In terms of architecture, we compare two variants of the coarse-supervised branch: (i) 'Seq' being $\mathcal{B} \to \mathcal{E} \to C \to \mathcal{L}_{CE}$; and (ii) 'Fork' being $\mathcal{B} \to \mathcal{E}_{layer\#1} \to C \to \mathcal{L}_{CE}$. In both cases the self-supervised contrastive branch remains the same: $\mathcal{B} \to \mathcal{E} \to \mathcal{L}_{cont}$. We also test the contribution of the angular normalization $\mathcal{A}$ in the self-supervised branch (between $\mathcal{E}$ and $\mathcal{L}_{cont}$). Of Seq and Fork, only Seq admits the possibility for angular normalization, as in Fork the classifier weights are of different dimensionality ($\text{dim}=d$) than the output of $\mathcal{E}$ ($\text{dim}=e$). Finally, we also evaluate the contribution of having a queue $\mathcal{Q}_i$ for each class ('Multi') vs. simply stacking all negative keys in a single shared queue $\mathcal{Q}$ ('Single'). The results of this ablation are presented in Table \ref{tab:arch_ablation}. Looking at all-way results we draw the following conclusions: (i) difference between Seq and Fork is small; (ii) without angular normalization single queue is better than multi-queue in all-way, likely because $\mathcal{L}_{cont}$ negative keys $k_{-}$ taken from the shared queue do not belong to the same class (as opposed to Multi with queue per class) reducing the $\mathcal{L}_{cont}$ drive to disperse same class elements in feature space; (iii) angular normalization gives a significant boost both to Single+Seq and to Multi+Seq; and (iv) Multi+Seq is best when combined with angular normalization forming \ours.

\begin{table}[t]
\begin{center}
\resizebox{\linewidth}{!}{%
\begin{tabular}{lccccc}
% \toprule
& MQ &  S/F  & $\mathcal{A}$ & {5-way}  &  {all-way}  \\
% & per class &  classifier arch. & norm. & 5-way  & all-way \\
\midrule
Single+Fork & \xmark & F  & \xmark  & 82.45           & 42.75         \\
Multi+Fork  & \cmark  & F  & \xmark  & 85.38           & 36.28          \\
Single+Seq & \xmark & S   & \xmark  & 79.44          & 41.55           \\
Multi+Seq & \cmark  & S   & \xmark  & 81.75          & 36.06           \\
Single+Seq+Angular & \xmark & S  & \cmark  & 82.1 & 42.62 \\
\midrule
\oursspace (ours) \scriptsize{Multi+Seq+Angular} & \cmark  & S & \cmark  & \textbf{89.23}         & \textbf{45.14 } \\
% Multi+Fork+A (pass)    & 89.73       & 43.73           \\
% Multi+Fork+A (distill) & \textbf{90.12 } & 44.58           \\
\bottomrule
\end{tabular}}
\caption{\textbf{Architecture ablations}. (1) \textbf{MQ}: Multi queue (queue per class, otherwise one queue for all classes \cite{he2019moco}) (2) \textbf{S/F}: Sequential or Fork architecture. (3) \textbf{$\mathcal{A}$}: Angular normalization component. 
% Performed on LIVING-17 dataset.
}
    \vspace{-25pt}
\label{tab:arch_ablation}
\end{center}
\end{table}

\vspace{-11pt}
\subsubsection{The effect of adding more shots}\label{sec:shots}
\begin{figure}
    % \centering
    % \includesvg{figures/5way.svg}
    % \includesvg{figures/allway.svg}
    % \hspace{-20pt}
    % \includegraphics[width=0.46\linewidth]{figures/5way_many_shot.pdf}
    % \hspace{-15pt}
    % \includegraphics[width=0.46\linewidth]{figures/allway_many_shot.pdf}
    \vspace{-10pt}
    \centering
    \includegraphics[width=0.9\linewidth]{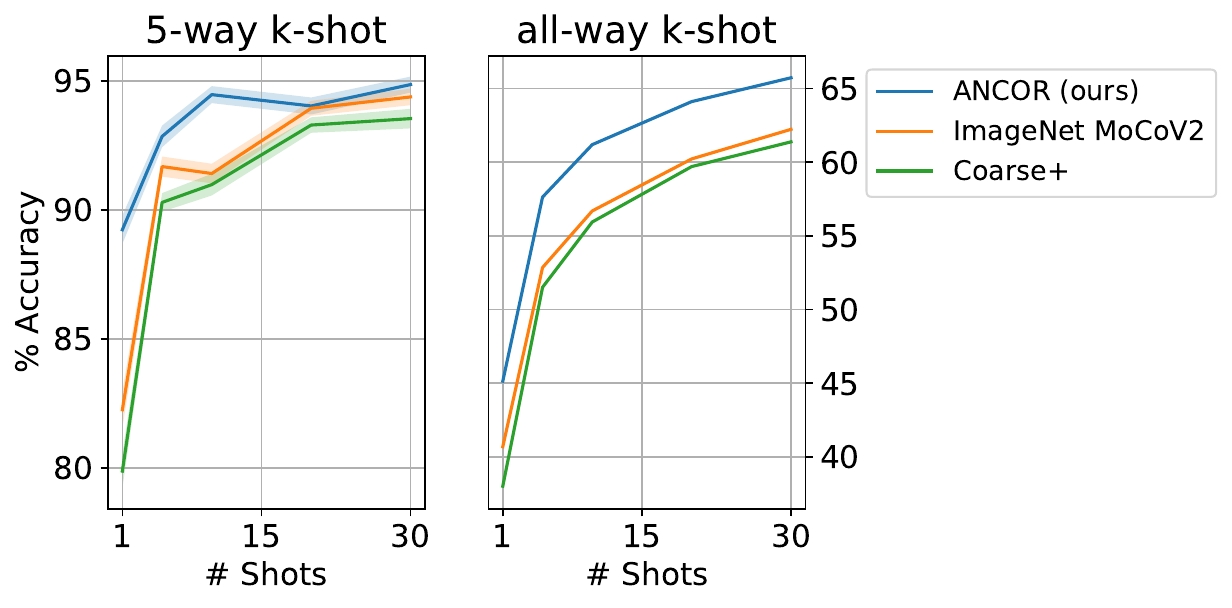}
    % \vspace{5pt}
    \caption{5-way and all-way results with increasing $k$ shots.}
    \vspace{-10pt}
    \label{fig:many_shot}
\end{figure}

\vspace{-0.1cm}
Here we evaluate the effect of adding more shots (number of support samples per class) to the few-shot episodes during testing. The results are shown in Fig. \ref{fig:many_shot} demonstrating \ours's consistent advantage (of about $+5\%$ accuracy in all-way) above the strongest of self-supervised (MocoV2-ImageNet) and supervised (Coarse+) baselines.

\begin{table}[htb]
\vspace{-7pt}
\begin{tabular}{lccc}
\toprule
             & all-way & intra-class   & all-way \\
             & (coarse) & (fine)   & (fine) \\
\midrule
Coarse       & 81.44 \scriptsize{$\pm$ 0.31}     & 37.03 \scriptsize{$\pm$ 0.53} & 33.83 \scriptsize{$\pm$ 0.1}    \\
Coarse$+$ & 50.83 \scriptsize{$\pm$ 0.32}     & 46.56 \scriptsize{$\pm$ 0.65} & 37.44 \scriptsize{$\pm$ 0.12}   \\
MoCoV2 & 27.36 \scriptsize{$\pm$ 0.23}     & 47.7 \scriptsize{$\pm$ 0.62} & 18.57 \scriptsize{$\pm$ 0.11}    \\
% MoCoV2-ImageNet & 56.77 \scriptsize{$\pm$ 0.3}     & \textbf{51.56 \scriptsize{$\pm$ 0.8}} & 40.29 \scriptsize{$\pm$ 0.14}    \\
\oursspace (ours)         & \textbf{84.25 \scriptsize{$\pm$ 0.31}}     & \textbf{48.77 \scriptsize{$\pm$ 0.71}} & \textbf{45.14 \scriptsize{$\pm$ 0.12}} \\ 
\bottomrule
\end{tabular}
\captionof{table}{Closer look on all-way (fine test labels) result breaking it into all-way with coarse test labels and intra-class-fine (fine classes of same random coarse class).
% Results for coarse classification and fine classification sub-tasks. A regular all-way task is intrinsically comprised of both of these together, and isolating them may help in understanding where our model has outperformed the baselines. Reported on LIVING-17.
}
    \vspace{-20pt}
\label{tab:coarse_vs_fine}
\end{table}
\vspace{-0.1cm}
\subsubsection{Closer look at the fine classes performance}\label{sec:where}
\vspace{-0.1cm}
Success in \ourstaskspace entails both not confusing the coarse classes as well as not confusing the fine sub-classes of each class. Therefore we ask ourselves, is the benefit of \oursspace coming from less coarse confusion or from less (intra-class) fine confusion? Looking at Table \ref{tab:coarse_vs_fine} we see that \oursspace has better all-way coarse accuracy (using coarse-only labels of the test set), as well as better intra-class fine accuracy evaluated by creating the few-shot episodes by random sampling a coarse class $y_i \in \mathcal{Y}_{coarse}$ and then generating a random episode from (all of) it's sub-classes $\{y_{i,1},...,y_{i,k_i}\}$. Consequently, \oursspace has the highest performance on the all-way fine test that we also saw in Tab. \ref{tab:breeds}. Interestingly, Coarse, Coarse$+$, and MoCoV2 results indicate there is a trade-off between all-way coarse and intra-class-fine performances, trade-off apparently topped by \ours.

\vspace{-10pt}
\subsubsection{Closer look at the features}
\begin{figure}[b]
    \vspace{-10pt}
    \centering
    % \hspace{-15pt}
    \includegraphics[width=0.9\linewidth]{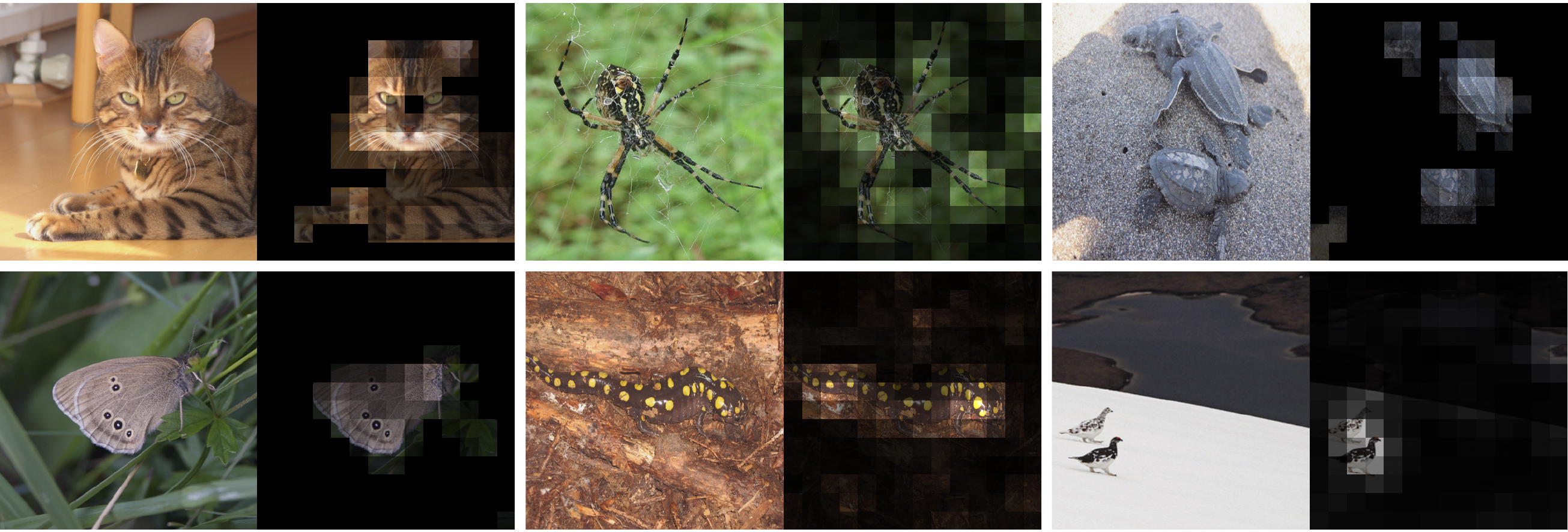}
    \caption{\oursspace encoder $\mathcal{B}$ last layer activations examples.}\label{fig:localization}
    \vspace{-10pt}
\end{figure}

\vspace{-0.1cm}
We further explore the feature space $\mathcal{F}$ learned by \oursspace using visualizations. Fig. \ref{fig:tsne} visualizes $\mathcal{F}$ on ENTITY-13 via tSNE and compares it to the tSNE plots of the feature space of the Coarse baseline, both on the coarse classes level and when digging into a random coarse class, qualitatively showing an advantage for \oursspace's feature space.
In addition, in figure \ref{fig:localization} we show a simple heatmap visualization of activations of the last convolutional layer of $\mathcal{B}$ obtained by dropping the GAP at the end of $\mathcal{B}$ and for each feature vector $f$ corresponding to a spatial coordinate in the resulting tensor computing the norm of the activation $||\mathcal{E}(f)||^2$. To obtain a higher resolution for this visualization we also increase the input image resolution by $\times2$. Surprisingly, despite the relatively small number of coarse classes (not benefiting spatial specificity), and the instance recognition nature of the self supervised objective (that could in principle use background pixels to discriminate instance images), the features learned by \oursspace trained $\mathcal{B}$ are remarkably good at localizing the object instances essentially ignoring the background. We verified this phenomena is stable and repeating in almost all localizations we examined and would be very interesting to explore in future studies.

\vspace{-10pt}
\subsubsection{Longer training}\label{sec:epochs}
\begin{figure}[t]
    % \centering
    % \includesvg{figures/5way.svg}
    % \includesvg{figures/allway.svg}
    % \hspace{-20pt}
    \vspace{-10pt}
    \centering
    \includegraphics[width=0.9\linewidth]{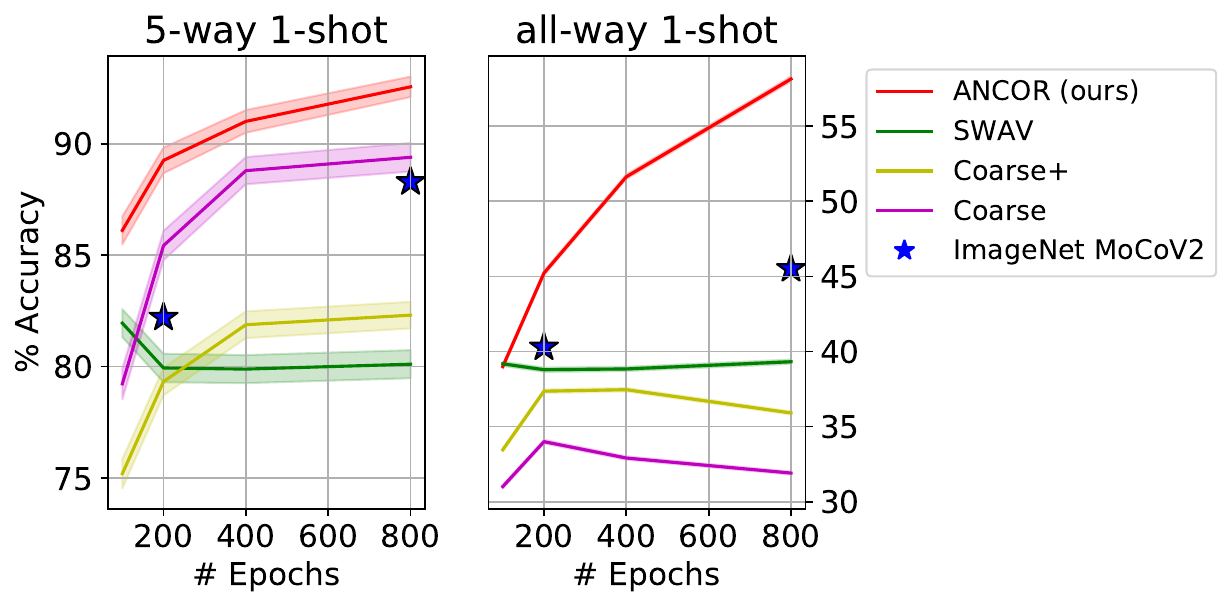}
    % \hspace{-15pt}
    % \includegraphics[width=0.46\linewidth]{figures/many_epochs_aw.pdf}
    % \includegraphics[width=0.6\textwidth]{figures/manyshot.pdf}
    \caption{Results for 5-way and all-way tests, reporting with an increasing number of epochs.}
    \vspace{-15pt}
    \label{fig:many_epochs}
\end{figure}

\vspace{-0.1cm}
In all above experiments, for fair comparison we used 200 epochs for training all models (\ours, baselines and upper bounds). In Fig. \ref{fig:many_epochs} we explore what happens if we train longer. As can be seen, there is still much to be gained from \oursspace with longer training. We attribute this positive effect to the contrastive component that is known to benefit from longer training \cite{Caron2020,chen2020mocov2}. Almost 15\% gains are observed for \oursspace when increasing the number of epochs from 200 to 800, and interestingly, \oursspace's 'all-way' gain above the baselines becomes larger with more epochs.
%
% \begin{figure}
%     % \hspace{-20pt}
%     \centering
%     \includegraphics[width=0.8\linewidth]{figures/many_epochs.pdf}
%     \caption{Accuracy on LIVING-17, for an extra amount of epochs.}
%     \label{fig:many_epochs}
% \end{figure}

%

% \begin{figure*}
% \centering
% \begin{minipage}{.4\textwidth}
% %   \centering
%   \input{tables/cifar100}
% %   \captionof{figure}{A figure}
% %   \label{fig:test1}
% \end{minipage}%
% \begin{minipage}{.6\textwidth}
% %   \centering
% %   \input{tables/coarse_classification}
% %   \captionof{figure}{Another figure}
% %   \label{fig:test2}
% \end{minipage}
% \end{figure*}
\vspace{-5pt}
\section{Summary and conclusions}\label{sec:conclusions}
\vspace{-5pt}
We have proposed the \ourstaskspace task focusing on situations when a few-shot model needs to adapt to much finer-grained unseen classes than the base classes used during its pre-training, including the challenging case when the unseen target classes are sub-classes of the base classes. We introduced the \oursspace approach for the \ourstaskspace task based on effective combination of inter-class supervised and intra-class self-supervised losses featuring a novel 'angular normalization' component inducing synergy between these (otherwise conflicting) losses. We have demonstrated the effectiveness of \oursspace on a variety of datasets showing: (i) promising results of \oursspace on the \ourstaskspace task also in the more challenging 'all-way' setting; (ii) that the proposed angular component is instrumental in the success of \ours; (iii) the advantages of \oursspace are preserved when adding more shots for the fine classes; (iv) \oursspace does best also in the challenging 'intra-class fine' scenario when all target classes belong to the same coarse class; (v) promising properties of \oursspace feature space including surprisingly good spatial attention to object instances; and (vi) \oursspace can be improved considerably with longer training, getting larger improvements on \ourstaskspace task than even leading self-supervised methods trained on more data (\cite{chen2020mocov2,Caron2020}). We hope that this work will serve as a good basis for future research into the exciting and challenging \ourstaskspace task further pushing its limits. Finally, we believe that the proposed angular normalization component is useful beyond the \ourstaskspace task for any situation involving supervised and contrastive self-supervised multi-tasking, and leave it to future works to explore its uses further.
\section{Appendix}
% \vspace{-0.2cm}
\subsection{Additional baselines}
\begin{table*}[h]
\resizebox{\textwidth}{!}{%
\begin{tabular}{l@{\hskip 12pt}c@{\hskip 6pt}c@{\hskip 12pt}c@{\hskip 6pt}c@{\hskip 12pt}c@{\hskip 6pt}c@{\hskip 12pt}c@{\hskip 6pt}c}
\toprule
 &
  \multicolumn{2}{c}{LIVING-17} &
  \multicolumn{2}{c}{NONLIVING-26} &
  \multicolumn{2}{c}{ENTITY-13} &
  \multicolumn{2}{c}{ENTITY-30} \\
Method              & 5-way        & all-way      & 5-way        & all-way      & 5-way        & all-way      & 5-way        & all-way      \\
\midrule
ensemble(Coarse+, MoCoV2)         & 74.07 \scriptsize{$\pm$ 0.67}  & 34.02 \scriptsize{$\pm$ 0.13} & 73.14 \scriptsize{$\pm$ 0.71} & 33.23 \scriptsize{$\pm$ 0.11} & 85.27 \scriptsize{$\pm$ 0.57}  & 36.39 \scriptsize{$\pm$ 0.08} & 84.22 \scriptsize{$\pm$ 0.6} & 34.56 \scriptsize{$\pm$ 0.08} \\
ensemble(Coarse, MoCoV2)         & 84.32 \scriptsize{$\pm$ 0.62}  & 36.39 \scriptsize{$\pm$ 0.13} & 79.95 \scriptsize{$\pm$ 0.65} & 34.31 \scriptsize{$\pm$ 0.11} & 85.97 \scriptsize{$\pm$ 0.59}  & 34.03 \scriptsize{$\pm$ 0.08} & 88.48 \scriptsize{$\pm$ 0.55} & 34.49 \scriptsize{$\pm$ 0.08} \\
cascade(Coarse+,MoCoV2)         & 74.1 \scriptsize{$\pm$ 0.66}  & 34.04 \scriptsize{$\pm$ 0.13} & 73.1 \scriptsize{$\pm$ 0.7} & 33.23 \scriptsize{$\pm$ 0.11} & 85.19 \scriptsize{$\pm$ 0.57}  & 36.38 \scriptsize{$\pm$ 0.08} & 84.33 \scriptsize{$\pm$ 0.6} & 34.57 \scriptsize{$\pm$ 0.08} \\
cascade(Coarse,MoCoV2)         & 88.27 \scriptsize{$\pm$ 0.63}  & 38.02 \scriptsize{$\pm$ 0.14} & 84.66 \scriptsize{$\pm$ 0.64} & 36.05 \scriptsize{$\pm$ 0.11} & 86.39 \scriptsize{$\pm$ 0.59}  & 35.46 \scriptsize{$\pm$ 0.07} & \textbf{90.02 \scriptsize{$\pm$ 0.55}} & 37.1 \scriptsize{$\pm$ 0.08} \\
concat(Coarse+,MoCoV2)         & 73.48 \scriptsize{$\pm$ 0.66}  & 33.4 \scriptsize{$\pm$ 0.13} & 71.49 \scriptsize{$\pm$ 0.71} & 31.28 \scriptsize{$\pm$ 0.11} & 84.26 \scriptsize{$\pm$ 0.59}  & 35.64 \scriptsize{$\pm$ 0.08} & 83.03 \scriptsize{$\pm$ 0.62} & 33.5 \scriptsize{$\pm$ 0.07} \\
concat(Coarse,MoCoV2)         & 83.45 \scriptsize{$\pm$ 0.63}  & 35.89 \scriptsize{$\pm$ 0.13} & 77.59 \scriptsize{$\pm$ 0.67} & 32.48 \scriptsize{$\pm$ 0.11} & 85.35 \scriptsize{$\pm$ 0.6}  & 33.89 \scriptsize{$\pm$ 0.07} & 86.83 \scriptsize{$\pm$ 0.57} & 33.82 \scriptsize{$\pm$ 0.07} \\
Supervised Contrastive \cite{Khosla2020}         & 86.49 \scriptsize{$\pm$ 0.67}  & 35.11 \scriptsize{$\pm$ 0.12} & 84.54 \scriptsize{$\pm$ 0.61} & 37.44 \scriptsize{$\pm$ 0.11} & 87.08 \scriptsize{$\pm$ 0.58}  & 28.57 \scriptsize{$\pm$ 0.15} & 88.86 \scriptsize{$\pm$ 0.52} & 33.67 \scriptsize{$\pm$ 0.17} \\
\midrule
\oursspace (ours) &
  \textbf{89.23 \scriptsize{$\pm$ 0.55}} &
  \textbf{45.14 \scriptsize{$\pm$ 0.12}} &
  \textbf{86.23 \scriptsize{$\pm$ 0.54}} &
  \textbf{43.10 \scriptsize{$\pm$ 0.11}} &
  \textbf{90.58 \scriptsize{$\pm$ 0.54}} &
  \textbf{42.29 \scriptsize{$\pm$ 0.08}} &
  88.12 \scriptsize{$\pm$ 0.54} &
  \textbf{41.79 \scriptsize{$\pm$ 0.08}} \\ 
 \bottomrule
\end{tabular}}
\caption{
\textbf{Additional baselines:} evaluated using BREEDS \cite{SanturkarS2020} with $1$-shot. The "Coarse", "Coarse+", and "MoCoV2" models used in ensemble and cascade combinations are as described in the paper. 
(1) \textbf{ensemble:} averaging the two models predictions (probabilities after softmax); (2) \textbf{cascade:} classify into the max-scoring coarse classes $\mathcal{Y}_{coarse}$ using the coarse-supervised model (with the linear classifier resulting from pre-training), and then use the self-supervised model (with the LR classifier, paper Sec. 3.3) for intra-class classification within the chosen coarse class (limiting the target classes to its sub-classes); (3) \textbf{concat:} concatenate the feature vectors from the two models and then apply few-shot classification as in the paper Sec. 3.3;
(4) Same as Coarse+ using the Supervised Contrastive loss \cite{Khosla2020} replacing CE.
}
    \vspace{-10pt}
\label{tab:extra_baselines}
\end{table*}
\begin{table*}[h!]
\begin{center}
% \resizebox{\linewidth}{!}{%
\begin{tabular}{lcccc}
\toprule
& Good &  Bad  & {5-way}  &  {all-way}  \\
% & per class &  classifier arch. & norm. & 5-way  & all-way \\
\midrule
Coarse$+$ &  & \cmark &  70.77 \scriptsize{$\pm$ 0.74} & 36.65 \scriptsize{$\pm$ 0.19}        \\
Coarse$+$  & \cmark  &   & 73.44 \scriptsize{$\pm$ 0.69}  & 43.13 \scriptsize{$\pm$ 0.22} \\
\midrule
\oursspace (ours) &  & \cmark   & 74.99  \scriptsize{$\pm$ 0.71}        & 40.69 \scriptsize{$\pm$ 0.20} \\
\oursspace (ours) & \cmark &   & \textbf{77.32 \scriptsize{$\pm$ 0.69}}          & \textbf{47.53 \scriptsize{$\pm$ 0.22}}           \\
\bottomrule
\end{tabular}
% }
\caption{\textbf{Sub-population Shift}. Two hand-crafted partitions of the LIVING-17 dataset, created by \cite{SanturkarS2020}, such that the test sub-classes are different from the (unlabeled) training sub-classes, yet share the common coarse classes. 'Good' and 'Bad' represent a less and more adversarial partitioning. Note that in practice these models train on half the data the models trained on LIVING-17 in the main paper have, due to the partitioning.
% Performed on LIVING-17 dataset.
}\label{tab:subpopulation_shift}
% \vspace{-25pt}
\end{center}
\end{table*}
We include $7$ additional baseline comparisons using the four BREEDS \cite{SanturkarS2020} datasets. These baselines were not included in the main paper due to lack of space. The comparisons are summarized in Table \ref{tab:extra_baselines}. We include several kinds of additional baselines covering different possible supervised + self-supervised combinations, as well as of supervised contrastive loss use: 
\begin{itemize}
    \item \textbf{ensemble (of Coarse/Coarse+ with MoCoV2):} averaging the two models predictions (probabilities after softmax)
    \item \textbf{cascade (of Coarse/Coarse+ with MoCoV2):} classify into the max-scoring coarse classes $\mathcal{Y}_{coarse}$ using the coarse-supervised model (with the linear classifier resulting from pre-training), and then use the self-supervised model (with the LR classifier, Section 3.3 in the paper) for intra-class classification within the chosen coarse class (limiting the target classes to the sub-classes of the predicted coarse class)
    \item \textbf{concat (of Coarse/Coarse+ with MoCoV2):} concatenating the features produces by the two models, and doing few-shot classification via learning the logistic regression (paper Section 3.3) on the resulting concatenated features.
    \item \textbf{Supervised Contrastive:} Training with coarse labels $\mathcal{Y}_{coarse}$ using the Supervised Contrastive loss \cite{Khosla2020} replacing CE.
\end{itemize}
The "Coarse", "Coarse+", and "MoCoV2" models used in ensemble, cascade, and concat combinations are as described in the paper. As can be seen from the table, in all (the more challenging) all-way experiments \oursspace maintains a significant advantage over the baselines, even ones combining (via an ensemble, a cascade, or a concat) separately trained coarse-supervised and contrastive self-supervised models, and thus also using twice more learn-able parameters than \ours. This demonstrates once again the importance of joint coarse-supervised and contrastive self-supervised training employed by \oursspace and facilitated by the proposed angular normalization component enhancing the synergy between these two objectives.

% \section{Code}
% We provide our code in "code.zip". The instructions for use are included in "README.md" inside the zip.

\begin{table*}[htb!]
\resizebox{\textwidth}{!}{
\begin{tabular}{l@{\hskip 12pt}c@{\hskip 6pt}c@{\hskip 12pt}c@{\hskip 6pt}c@{\hskip 12pt}c@{\hskip 6pt}c@{\hskip 12pt}c@{\hskip 6pt}c}
\toprule
 &
  \multicolumn{2}{c}{LIVING-17} &
  \multicolumn{2}{c}{NONLIVING-26} &
  \multicolumn{2}{c}{ENTITY-13} &
  \multicolumn{2}{c}{ENTITY-30} \\
Method              & 5-way        & all-way      & 5-way        & all-way      & 5-way        & all-way      & 5-way        & all-way      \\
\midrule

Fine (upper-bound) & 91.94 \scriptsize{$\pm$ 0.44}  & 64.66 \scriptsize{$\pm$ 0.17} & 87.68 \scriptsize{$\pm$ 0.50} & 54.42 \scriptsize{$\pm$ 0.13} & 94.22 \scriptsize{$\pm$ 0.34} & 64.30 \scriptsize{$\pm$ 0.09} & 93.11 \scriptsize{$\pm$ 0.38} & 61.70 \scriptsize{$\pm$ 0.09} \\
Fine$+$ (upper-bound)    & 90.25 \scriptsize{$\pm$ 0.48} & 63.54 \scriptsize{$\pm$ 0.17} & 86.27 \scriptsize{$\pm$ 0.51} & 53.16 \scriptsize{$\pm$ 0.14} & 91.99 \scriptsize{$\pm$ 0.40} & 59.43 \scriptsize{$\pm$ 0.09} & 91.03 \scriptsize{$\pm$ 0.43} & 57.48 \scriptsize{$\pm$ 0.09} \\
\midrule
MoCoV2         & 81.45 \scriptsize{$\pm$ 0.61}  & 46.65 \scriptsize{$\pm$ 0.16} & 78.33 \scriptsize{$\pm$ 0.65} & 42.08 \scriptsize{$\pm$ 0.12} & 87.30 \scriptsize{$\pm$ 0.53}  & 48.97 \scriptsize{$\pm$ 0.08} & 86.51 \scriptsize{$\pm$ 0.56} & 46.77 \scriptsize{$\pm$ 0.09} \\
MoCoV2-ImageNet \cite{chen2020mocov2} & 89.27 \scriptsize{$\pm$ 0.57} & 51.60 \scriptsize{$\pm$ 0.15} & 82.22 \scriptsize{$\pm$ 0.66} & 43.32 \scriptsize{$\pm$ 0.12} & 88.30  \scriptsize{$\pm$ 0.55}  & 45.52 \scriptsize{$\pm$ 0.08} & 87.18 \scriptsize{$\pm$ 0.58} & 42.23 \scriptsize{$\pm$ 0.08} \\
SWAV-ImageNet \cite{Caron2020} & 80.11 \scriptsize{$\pm$ 0.63} & 39.30 \scriptsize{$\pm$ 0.14} & 73.43 \scriptsize{$\pm$ 0.67} & 33.06 \scriptsize{$\pm$ 0.11} & 79.58 \scriptsize{$\pm$ 0.62} & 33.36 \scriptsize{$\pm$ 0.07} & 78.89 \scriptsize{$\pm$ 0.64} & 31.16 \scriptsize{$\pm$ 0.07} \\
Coarse        & 89.04 \scriptsize{$\pm$ 0.63} & 29.06 \scriptsize{$\pm$ 0.23}  & 84.72 \scriptsize{$\pm$ 0.63} & 27.99 \scriptsize{$\pm$ 0.18} & 82.66 \scriptsize{$\pm$ 0.75} & 11.24 \scriptsize{$\pm$ 0.08} & 90.09 \scriptsize{$\pm$ 0.59} & 20.63 \scriptsize{$\pm$ 0.12} \\
Coarse$+$  & 89.41 \scriptsize{$\pm$ 0.61} & 33.07 \scriptsize{$\pm$ 0.23} & 84.69 \scriptsize{$\pm$ 0.59} & 32.07 \scriptsize{$\pm$ 0.20}  & 85.23 \scriptsize{$\pm$ 0.63} & 22.85 \scriptsize{$\pm$ 0.13} & 88.43 \scriptsize{$\pm$ 0.55} & 28.33 \scriptsize{$\pm$ 0.15} \\
\midrule
\oursspace (ours) &
  \textbf{92.59 \scriptsize{$\pm$ 0.47}} &
  \textbf{58.15 \scriptsize{$\pm$ 0.16}} &
  \textbf{88.25 \scriptsize{$\pm$ 0.52}} &
  \textbf{49.38 \scriptsize{$\pm$ 0.13}} &
  \textbf{92.04 \scriptsize{$\pm$ 0.44}} &
  \textbf{50.72 \scriptsize{$\pm$ 0.09}} &
  \textbf{92.13 \scriptsize{$\pm$ 0.44}} &
  \textbf{50.85 \scriptsize{$\pm$ 0.09}} \\ 
 \bottomrule
\end{tabular}}
\caption{Results for different baselines on the four BREEDS datasets. Every model was trained for 800 epochs.}
\label{tab:breeds_800}
\end{table*}
\subsection{Sub-population Shift}

The 'sub-population shift' benchmark was proposed in \cite{SanturkarS2020} intended to evaluate how classification performance is affected when the train classes and test classes consist of different (non-overlapping set of) sub-classes (eg. `Dog` class in training consist of samples of 'Bloodhound' and 'Pekinese', while the test dogs are the 'Great Pyrenese' and the 'papillon'). For this purpose they propose two hand-crafted partitions for each of their datasets: named 'good' and 'bad', which represent a less and a more adversarial partitioning respectively. We leverage this to create a task that is on one hand allows having the same coarse classes in training and in testing (as in the task in Section 4.4.1), while still having the sub-classes of those coarse classes non-overlapping (different) between the train and the test (as in the task in Section in 4.4.2). In other words, in this scenario the test sub-classes are completely different from the training ones, and yet they share the same parent coarse classes. Our evaluation on this task, provided in Table \ref{tab:subpopulation_shift}, shows that despite this challenging setting, \oursspace significantly outperforms the strongest baseline by $>4\%$ in the all-way test.

\subsection{Additional results for 800-epoch training}
In this section we further examine the effect of longer training longer on the performance. As can be seen in Table \ref{tab:breeds_800}, following longer 800 epoch training \oursspace obtains significant gains in all the experimental settings. The most noticeable gains are in the all-way tests, where we observe that the gap above the baselines grows with longer training. We attribute this improvement to the contrastive component that is known to benefit from longer training \cite{Caron2020,chen2020mocov2}. Interestingly, with longer training the coarse baseline models have gained accuracy in the 5-way test, but lost accuracy in the all-way test (compared to the 200 epochs performance). This supports our hypothesis that the coarse-classes supervised objective encourages reducing intra-class variation, and as such, with the longer training, tends to decresae the distinction between fine-sub classes loosing their discriminability within the coarse class. This again underlines the merits of our \oursspace approach that retains and enhances the fine sub-classes discrimnability thus significantly benefiting from the longer training regime.

% doesn't reach this state of overfitting after 800 epochs further highlights the usefulness of this method. Finally, these results also demonstrates what we consider to be a down-side to the 5-way test: Since the number of coarse classes is $\gg5$, the chance of choosing 5 ways that have different coarse classes is high. This causes the 5-way test to have heavy bias towards coarse classification. As can be observed in Tab.~\ref{tab:breeds_800}, the gap in performance between coarse and fine models in 5-way are small. However, in a more complete setting (i.e. all-way), there is a huge gap in performance among them.

\begin{figure*}[htb!]
    \vspace{-10pt}
    \centering
    % \hspace{-15pt}
    \includegraphics[width=1.0\linewidth]{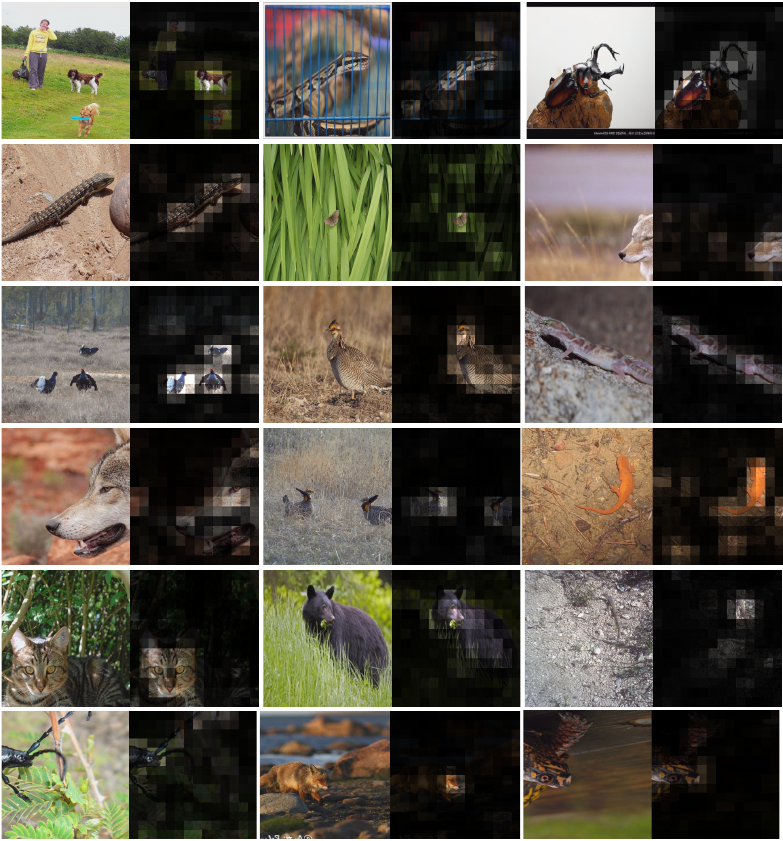}
    \caption{Additional examples of \oursspace encoder $\mathcal{B}$ last layer activations.}\label{fig:more_activations}
    \vspace{-10pt}
\end{figure*}

\subsection{Additional examples of \oursspace encoder $\mathcal{B}$ last layer activations}
Additional examples of \oursspace encoder $\mathcal{B}$ last layer activations are provided in Figure \ref{fig:more_activations}, again illustrating an interesting attention to objects learned by \oursspace despite not being provided with any location supervision during training.

\noindent\textbf{Acknowledgments}
This material is based upon work supported by the Defense Advanced Research Projects Agency (DARPA) under Contract No. FA8750-19-C-1001. Any opinions, findings and conclusions or recommendations expressed in this material are those of the author(s) and do not necessarily reflect the views of DARPA. Raja Giryes was supported by ERC-StG grant no. 757497 (SPADE).

{\small
\bibliographystyle{ieee_fullname}
\bibliography{references}
}

\end{document}